%% file: main.tex
\documentclass[10pt]{article} 
\usepackage[table,x11names]{xcolor}
\usepackage[accepted]{tmlr}

\input{math_commands.tex}

\usepackage{hyperref}
\usepackage{url}
\usepackage{algorithm}
\usepackage{amsfonts}
\usepackage{graphicx}
\usepackage{booktabs}
\usepackage{array}
\usepackage{subcaption}
\usepackage{amsmath}
\usepackage{hyperref}
\usepackage{cleveref}
\usepackage{algpseudocode}

\title{ADAPT to Robustify Prompt Tuning Vision Transformers}

\author{\name Masih Eskandar \email eskandar.m@northeastern.edu \\
      \addr Department of Electrical \& Computer Engineering\\
      Northeastern University
      \AND
      \name Tooba Imtiaz \email imtiaz.t@northeastern.edu \\
      \addr Department of Electrical \& Computer Engineering\\
      Northeastern University
      \AND
      \name Zifeng Wang \email zifengwang@ece.neu.edu\\
      \addr Department of Electrical \& Computer Engineering\\
      Northeastern University
      \AND
      Jennifer Dy \email jdy@ece.neu.edu\\
      \addr Department of Electrical \& Computer Engineering\\
      Northeastern University
      }

\def\month{01}  
\def\year{2025} 
\def\openreview{\url{https://openreview.net/forum?id=bZzXgheUSD}} 

\begin{document}

\maketitle
\begin{abstract}
The performance of deep models, including Vision Transformers, is known to be vulnerable to adversarial attacks. Many existing defenses against these attacks, such as adversarial training, rely on full-model fine-tuning to induce robustness in the models. These defenses require storing a copy of the entire model, that can have billions of parameters, for each task. At the same time, parameter-efficient prompt tuning is used to adapt large transformer-based models to downstream tasks without the need to save large copies. In this paper, we examine parameter-efficient prompt tuning of Vision Transformers for downstream tasks under the lens of robustness. We show that previous adversarial defense methods, when applied to the prompt tuning paradigm, suffer from \textit{gradient obfuscation} and are vulnerable to adaptive attacks. We introduce ADAPT, a novel framework for performing adaptive adversarial training in the prompt tuning paradigm.  Our method achieves competitive robust accuracy of $\sim40\%$ w.r.t. SOTA robustness methods using full-model fine-tuning, by tuning only $\sim 1\%$ of the number of parameters.
\end{abstract}

%

\section{Introduction}
\label{sec:intro}


Despite their success, deep neural networks, including the popular ViT architecture, have been shown to be vulnerable to adversarial attacks \citep{szegedy2013intriguing, mahmood2021robustness}. 
These attacks are modifications to input images, which are generally imperceptible to humans but can drastically change the prediction of a neural network. Many countermeasures have been introduced to induce robustness against such attacks, one common approach being adversarial training \citep{madry2018towards}, where the input is augmented with said perturbations during training. While originally introduced for CNNs, adversarial training has been extended to ViTs in the full-model fine-tuning paradigm \citep{mo2022adversarial}. However, adversarial training and similar methods are computationally expensive and difficult to perform for large datasets as it requires performing multi-step evaluation of adversarial examples. Hence ViTs are generally pre-trained without such considerations.

Transformer-based language models \citep{brown2020language} which can consist of billions of parameters require pretraining on massive amounts of data. To fine-tune such models for different tasks, one would have to save a separate copy of the weights per task, making it highly inefficient in terms of parameters. However, it is shown that these models can be adapted to downstream tasks with fewer parameters, for example using prompt design \citep{brown2020language} or prompt tuning (PT) \citep{lester2021power}. In prompt tuning, optimizable tokens are prepended to the model input in order to leverage the pretraining knowledge and adapt to the downstream task without changing the weights of the model. This allows for a lightweight adaptation of these large models. ViTs can benefit from the same technique \citep{jia2022visual}, especially as their parameters continue to increase \citep{dehghani2023scaling}. 


It is therefore important to explore the utility of prompt tuning for the adversarial robustness of ViTs. In this paper, we show that existing adversarial defense methods, when applied to the prompt tuning paradigm, suffer from \textit{gradient obfuscation} \citep{athalye2018obfuscated}. This means the gradients of the model are not useful for obtaining adversarial perturbations, and training using existing methods will lead to a \textit{false sense of security}. We demonstrate this phenomenon empirically using single-step attacks, and then design an adaptive attack to expose the vulnerability of existing methods. Finally, we propose a framework for \textbf{AD}aptive \textbf{A}dversarial \textbf{P}rompt \textbf{T}uning (ADAPT) to overcome this vulnerability. We empirically demonstrate the superior robustness of our method in multiple settings.

\begin{figure*}
   \centering
     \begin{subfigure}[b]{0.45\textwidth}
         \centering
         \includegraphics[width=\textwidth]{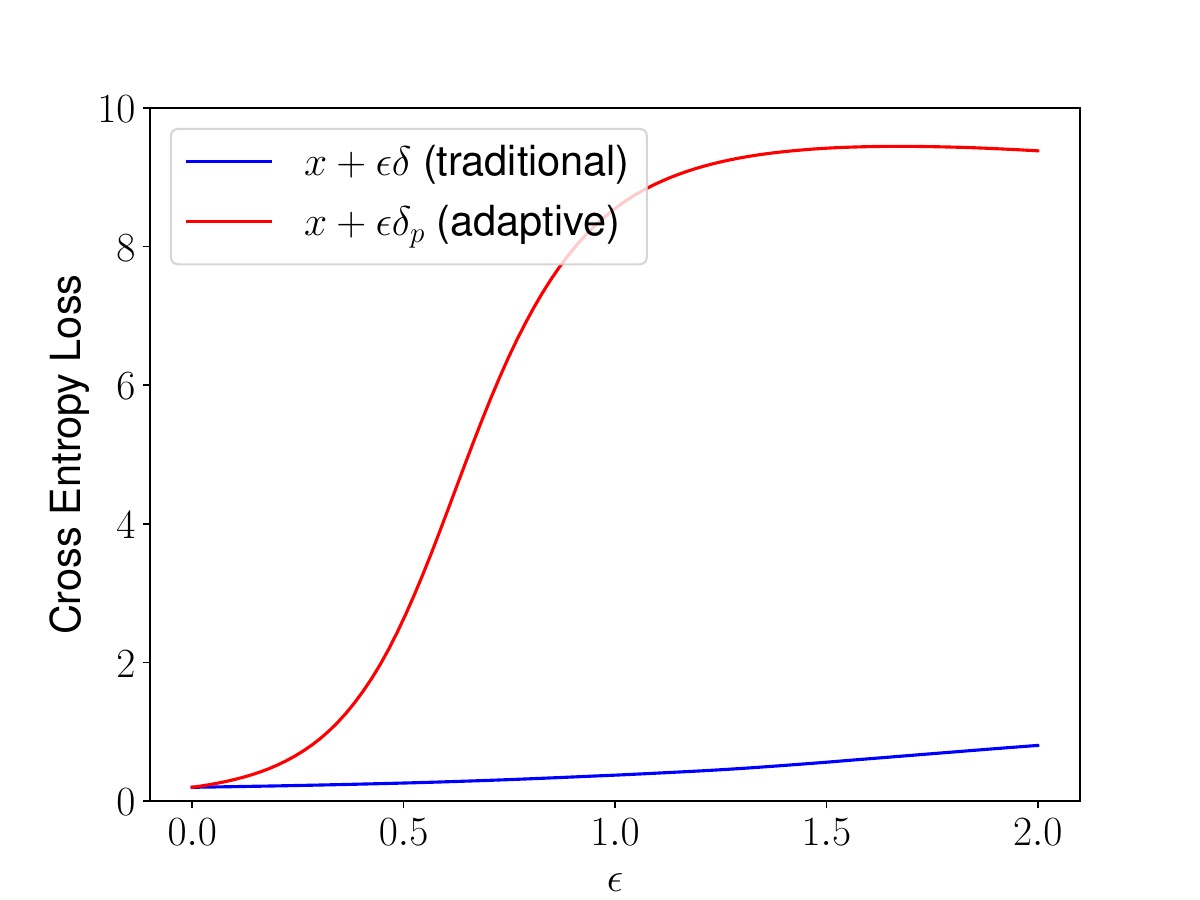}
         \caption{Adversarial Training + Prompt Tuning}
         \label{fig:pt}
     \end{subfigure}
     \begin{subfigure}[b]{0.45\textwidth}
         \centering
         \includegraphics[width=\textwidth]{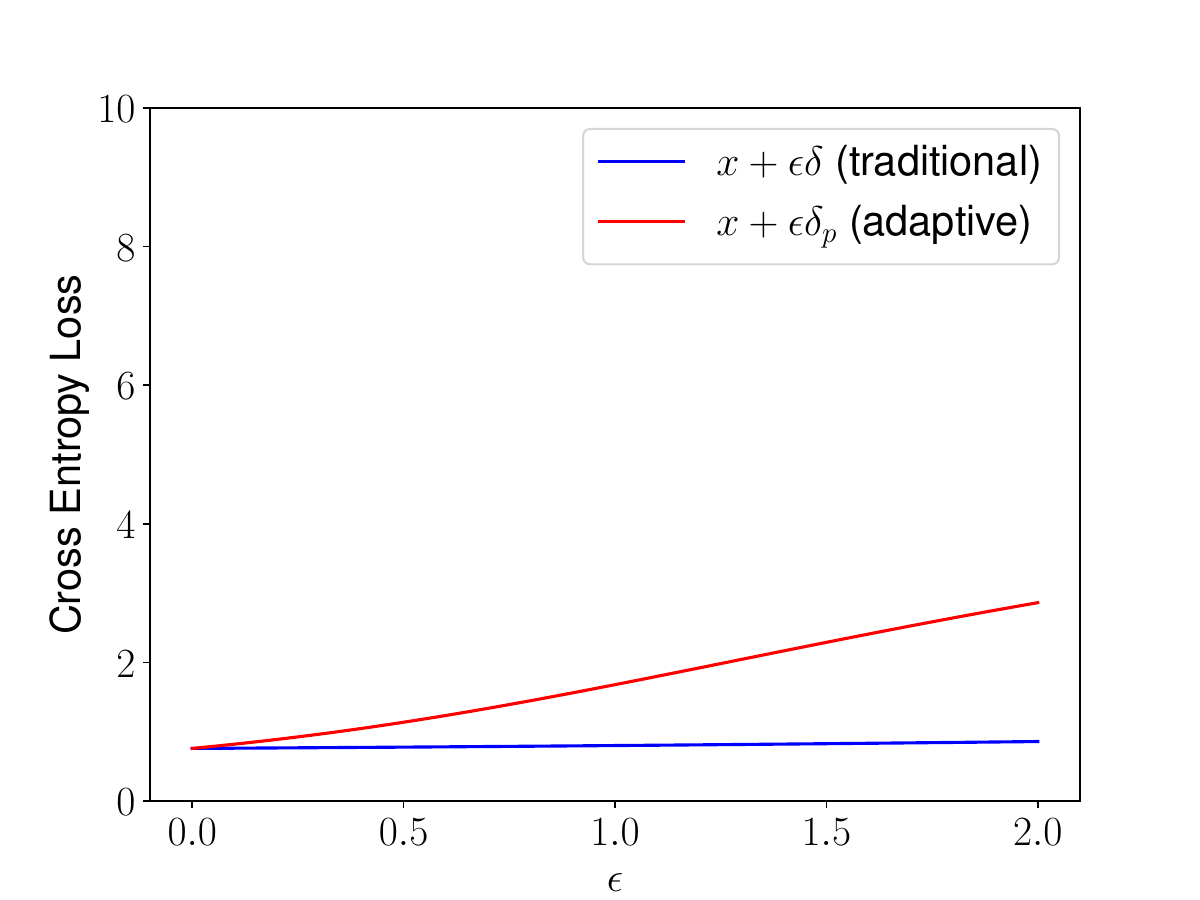}
         \caption{$\text{ADAPT}_{CE}$ + Prompt Tuning}
         \label{fig:ours}
     \end{subfigure}
     
     \caption{  \textbf{Existing methods do not exhibit robustness to adaptive attacks.} 
     Comparison of the cross entropy loss of a random batch along two adversarial perturbation directions. 
     Left (a) depicts the loss values of a prompt trained with traditional adversarial training and right (b) shows the loss values of a prompt trained with $\text{ADAPT}_{CE}$. Adversarial Training + Prompt Tuning does not exhibit a significant increase in the loss under the traditional PGD direction, but shows significant vulnerability to adaptive attacks. In contrast, the proposed method $\text{ADAPT}_{CE}$, exhibits robustness to both perturbation directions.}
     \label{fig:losses}
\end{figure*}

To the best of our knowledge, we are the first to study this issue extensively. A summary of our contributions is as follows:
\begin{itemize}
    \item We investigate the adversarial robustness of the prompt tuning paradigm for ViTs. We show that the existing methods, when applied to the prompt tuning scenario, suffer from gradient obfuscation \citep{athalye2018obfuscated}.
    \item We design an adaptive attack accordingly and show that existing methods are significantly vulnerable to our attack, resulting in a drop in their robust accuracy to approximately $1\%$ (See \cref{fig:losses}). 
    \item We propose a novel loss function that emphasizes conditioning on the prompts to craft adversarial examples during training. We quantitatively demonstrate the superiority of our method (ADAPT), in achieving robustness compared to existing methods.
\end{itemize}
Our implementation is available on Github\footnote{\url{https://github.com/Gnomy17/robust_prompt}}.

\section{Related Works}
\paragraph{Vision Transformers.}
Inspired by the groundbreaking success of Transformers for Natural Language Processing (NLP) tasks \citep{vaswani2017attention, kenton2019bert, openai2023gpt4}, Vision Transformers (ViTs) \citep{dosovitskiy2020image} are an adaptation of the architecture for computer vision tasks. Similar to the `tokens' fed to NLP transformers, here the input images are split into fixed-size patches, which are then embedded and fed to the transformer layers of the ViT. ViTs are powerful architectures and achieve state-of-the-art performance on several downstream vision tasks
\citep{pmlr-v139-touvron21a, caron2021emerging, he2022masked}.

\paragraph{Prompting.} This technique was introduced in the NLP literature to eliminate the need to fine-tune large pre-trained language models for downstream tasks \citep{liu2023pre}. Initially, Prompting was used to fine-tune the behavior of language models by pre-pending a set of fixed \textit{prompt templates} to the input text. Intuitively, prompting reformulates the downstream task so that it closely resembles the task that a frozen model has been pre-trained to solve. For example, \cite{gao2020making} use prompting for few-shot learning of language models. Related to this work, \cite{raman2023model} investigate the effects of adversarial robustness of prompting language models. However, these works are fundamentally specific to the NLP literature, as they require pre-determined prompts, and investigate the problem in the text input domain which is discrete. The extension of fixed prompt templates to the vision space, where inputs are continuous is non-trivial. To that effect, prompt tuning was introduced \citep{lester2021power} which uses continuous tunable vectors in the embedding space as prompts. Prompt-tuning, \textit{i.e.}, training the prompts for downstream tasks, has been extended to ViTs and is highly effective for domain generalization \citep{zheng2022prompt}, continual learning \citep{wang2022learning}, and several other vision tasks.

\paragraph{Adversarial Robustness.}
Adversarial Training (AT) was first introduced for CNNs and was thoroughly investigated \citep{szegedy2013intriguing, goodfellow2014explaining, madry2018towards} with several variations \citep{buckman2018thermometer, zhang2019theoretically, zhang2020attacks}. ViTs have also been shown to be vulnerable to adversarial attacks \citep{mahmood2021robustness}.
However, AT and its variants are computationally expensive due to requiring multiple forward and backward passes to obtain adversarial examples to train with. This makes them difficult to use for pre-training  ViTs. Therefore, adversarial defense methods are typically only adopted during fine-tuning for downstream tasks \citep{mo2022adversarial}. 

\paragraph{Robust Prompts.}
Prompt tuning has been employed for enhancing the robustness of NLP transformers \citep{yang2022prompting}, which deal with discrete inputs. Extending the same robustness to the continuous input space for ViTs is non-trivial and has not been explored in the literature.

There are other works which explore robustness of visual prompt tuning \citep{chen2023visual}. However, they are not tailored to the transformer paradigm. Moreover, their number of tuned parameters grows with the number of classes in the dataset.
Visual prompting or adversarial reprogramming was proposed before the advent of prompt tuning for adapting CNNs to downstream tasks \citep{elsayed2018adversarial}. As such, we are the first to investigate robust prompting for vision transformers.

\section{Notations and Preliminaries}
\label{sec:preleminaries}

\subsection{Vision Transformer}
\label{subsec:vit}

The Vision Transformer \citep{dosovitskiy2020image} has proven to be quite effective at performing various computer vision tasks. For our purposes, we consider the classification task.
A transformer $f_{\Phi, \Psi}$ as a classifier consists of a feature extractor $\Phi$ and a linear classifier $\Psi$ operating on the extracted features. That is, 
\begin{align}
    \Phi : \mathcal{X} \rightarrow \mathbb{R}^d, \Psi: {\mathbb{R}}^d \rightarrow {\mathbb{R}}^{|\mathcal{Y}|}\\
    f_{\Phi, \Psi}(x) = softmax(\Psi(\Phi(x))) 
\end{align}
where $\mathcal{X}$ is the input space, $\mathcal{Y}$ the label space, $|\mathcal{Y}|$ is the number of classes, and $d$ is the hidden dimension of the transformer (which stays the same throughout its layers).
As the transformer architecture operates on inputs that are sequences of high-dimensional vectors, we consider the input space of the transformer to be the space of all $d$-dimensional (the hidden dimension of the transformer) sequences of arbitrary length $k$:
\begin{equation}
\label{eq:inputspace}
    \mathcal{X} := \underset{\forall{k} \in \mathbb{N}}{\bigcup} \mathbb{R}^{d\times k}
\end{equation} 
As such, to transfer images from the data distribution support, i.e. $\mathbb{R}^{h\times w \times c}$, to the transformer input space $\mathcal{X}$, the image is divided into $n$ non-overlapping patches and then transformed via a linear transformation or through a convolutional neural network (CNN). Formally, we have 
\begin{equation}
    T: \mathbb{R}^{h\times w \times c} \rightarrow \mathbb{R}^{d\times n}
\end{equation}

\noindent as the embedding function. For the sake of brevity, except in sections where the embedding function is being discussed directly, we omit $T$ from our notation and assume that inputs are being passed through it.

\subsection{Prompt Tuning}

With the advent of transformers and large language models with billions of parameters pre-trained on extremely large amounts of data, it is desirable to tune these models for downstream tasks at a minimal cost. 
While full-model fine-tuning is an effective option, it requires training and storing an entirely new set of model weights for each downstream task, which becomes inefficient in terms of memory as both the number of tasks and the number of model parameters grow. A simple yet effective solution to this is prompt tuning \citep{lester2021power}, which proceeds by simply appending tunable prompt tokens to the transformer input space. Depending on the model size and number of tokens, it can yield comparable performance to full-model fine-tuning.
This allows us to adapt to downstream tasks without having to store copies of extremely large models, and instead only storing parameter-efficient prompts.

Formally, a pre-trained Transformer $f_{\Phi^*, \Psi^*}$ is obtained by training on large amounts of data, \textit{i.e.},
\begin{equation}
     \Phi^*, \Psi^* = \underset{\Phi, \Psi}{\arg\min}\ \mathbb{E}_{x,y \sim P} \mathcal{L}(f_{\Phi, \Psi}(x), y).
\end{equation}
For supervised learning, this can be solved using Empirical Risk Minimisation (ERM), where $P$ is a data distribution used for pre-training, e.g. ImageNet, and $\mathcal{L}$ is the loss function used for training. 
The goal is then to adapt to a downstream data distribution $D$ given $\Phi^*$.

Full-model fine-tuning (which we will refer to as Fine-Tuning (FT)) seeks to find some $\hat{\Phi}, \hat{\Psi}$ to minimize the expected loss $\mathbb{E}_{x,y \sim D}\ \mathcal{L}(f_{\hat{\Phi}, \hat{\Psi}}(x), y)$ by using $\Phi^*$ as a starting point for $\hat{\Phi}$.
Prompt tuning, instead, seeks to solve the following optimization problem
\begin{equation}
        \hat{\theta}_{p}, \hat{\Psi} = \underset{\theta_{p}, \Psi}{argmin}\ \mathbb{E}_{x,y \sim D}\ \mathcal{L}(f(\theta_p,x), y)
\end{equation}
Prompt tuning can be performed in multiple ways. In the case of simple Prompt Tuning (PT), a set of prompts is prepended to the input layer of the transformer. A prompted ViT $f(\theta_p,x)$ is defined as follows
\begin{align}
    f(\theta_{p},x) &:= f_{\Phi^*, {\Psi}}([\theta_{p},x])\\
    \theta_{p} &\in \mathbb{R}^{d\times m}
\end{align}
where $[.,.]$ is the concatenation operator, $d$ is the hidden dimension of the transformer, and $m$ is the prompt length (recall that the input space of the transformer is arbitrary length sequences). 
Another way to perform prompt tuning is Prefix tuning \citep{li2021prefix} or PT-v2 tuning (PT2) \citep{liu2021p}, where similar to PT, the backbone weights are kept unchanged. However, a set of prompt tokens are instead directly prepended to the input of \textit{every} layer of the transformer, by only prepending to the key and value matrices of the self-attention mechanism (while keeping the same number of tokens per layer). That is, $f(\theta_p,x)$ in the case of PT2 is as follows
\begin{align}
    f(\theta_{p},x) &:= f_{[\theta_p,\Phi^*], {\Psi}}(x)\\
    \theta_{p} &\in \mathbb{R}^{L \times d\times m}
\end{align}
where $[\theta_p; \Phi^*]$ is shorthand for prepending tokens to the input of every layer of the frozen feature extractor $\Phi^*$, $L$ is the number of transformer layers and $m$ is the prompt length, \textit{i.e.} the number of tokens tuned per layer.

\subsection{Adversarial Robustness}
Given an arbitrary classifier $f: \mathcal{X} \rightarrow \mathcal{Y}$, input $x \in \mathcal{X}$ with the true label $y \in \mathcal{Y}$, an adversary seeks to find some \textit{imperceptible} perturbation $\delta$ with the same dimensionality as $x$, such that $x + \delta \in \mathcal{X}$ and $f(x + \delta) \neq y$. To achieve such imperceptibility, $\delta$ is typically constrained w.r.t. an $\ell_q$ norm, the most popular choices being $q=2$ and $q=\infty$.

An untargetted adversarial attack for a classifier $f$ bounded in $\ell_q$ norm by $\epsilon$ can be defined in the following way
\begin{align}
\begin{split}
    \label{eq:advexn}
    x' = x + \underset{\|\delta\|_{q} \leq \epsilon }{argmax}\ \mathcal{L}(f(x + \delta), y)
\end{split}
\end{align}

The adversarial example defined above is an optimization problem that is typically solved using methods such as Projected Gradient Descent (PGD) \citep{madry2018towards}. As deep learning models are highly susceptible to such norm-bounded perturbations \citep{szegedy2013intriguing}, it is crucial to defend against these attacks. However, it is important to note that PGD assumes full knowledge of the model parameters and inference algorithm, i.e. a \textit{white-box} setting, for the adversary. Adversarial robustness can also be investigated through the lens of adversaries with limited knowledge, namely the \textit{black-box} \citep{ilyas2018black}, and \textit{grey-box} \citep{xu2021grey} settings. In this work, we focus on the white-box scenario.

One widely used defense method against adversarial attacks is Adversarial Training \citep{madry2018towards}. If $f_\theta$ is a classifier parameterized by $\theta$, AT tries to solve the following minimax optimization problem:
\begin{equation}
\label{eq:at}
    \underset{\theta}{min}\ \ \mathbb{E}_{x,y \sim D}\ \biggr[\underset{\|\delta\|_{q} \leq \epsilon }{max}\ \mathcal{L}(f_\theta(x+\delta), y)\biggr]
\end{equation}
Practically, this objective trains the model on adversarially perturbed inputs with untargeted attacks.

\subsubsection{Obfuscated Gradients.}

An adversarial defense method suffers from \textit{gradient masking} or \textit{obfuscated gradients} \citep{athalye2018obfuscated} when it does not have \textit{"useful gradients"} for generating adversarial examples.
Typically, there are three main causes for gradient obfuscation: exploding and vanishing gradients, stochastic gradients, and shattered gradients. We show that traditional adversarial defense methods in the prompt tuning paradigm suffer from shattered gradients. This means that the gradient of the model w.r.t. the input is incorrect and does not properly maximize the classification loss.
To validate this claim, we use the observation of \citep{athalye2018obfuscated} that in the presence of {obfuscated gradients}, single-step attacks are typically more effective than their multi-step counterparts. 
\section{Robust Prompt Tuning}
\label{sec:robust}

In this section we introduce ADAPT, a novel framework for robustifying the parameter efficient prompt tuning of ViTs. We design an adaptive attack tailored to the prompt tuning paradigm by conditioning on the prompts during each gradient step. Using our adaptive attack formulation, we propose a novel loss function for adaptive robust training of prompts while fine tuning for downstream tasks. 


\subsection{Adaptive Adversarial Attack} \label{sec:adaptiveadvatk}
In the prompt tuning scenario, $\nabla_x f(x)$ does not necessarily equal $\nabla_x f(\theta_p,x)$ . Therefore, the existence of the prompt may lead to a form of {gradient obfuscation} \citep{athalye2018obfuscated}, \textit{giving us a false sense of security} (See \cref{fig:pt}). In the experiment section , we demonstrate this effect by designing an adaptive attack and empirically show that while previous adversarial training methods may be robust against traditional attacks, they are vulnerable to our adaptive attack which reduces their robust accuracy to nearly zero.

In the prompt tuning paradigm, we reformulate the optimization objective for finding an adversarial example for classifier $f$ as follows
\begin{align}
\label{eq:adpatk}
    x_p' = x + \underset{\|\delta_p\|_q \leq \epsilon}{argmax} \; \mathcal{L}_{CE} (f(\theta_p, x + \delta_p), y)
\end{align}
We can solve this optimization problem using Projected Gradient Descent \citep{madry2018towards} similar to \cref{eq:advexn}. Here, the key difference lies in conditioning on the prompt during each gradient step. 

\subsection{Prompt Adversarial Training}

To perform adversarially robust prompt tuning, conditioning on the prompt is crucial for both generating the adversarial examples, as well as training the prompt itself.

As such, we propose a loss function for \textbf{AD}aptive \textbf{A}dversarial \textbf{P}rompt \textbf{T}uning (ADAPT) 
\begin{align}
    \mathcal{L}_{ADAPT}= \mathcal{L}_p (f(\theta_p, x), y) + \lambda \mathcal{L}_{adv}
\label{loss:ADAPT}
\end{align}
where $\mathcal{L}_p$ is the standard prompt tuning loss for any downstream task, and $\mathcal{L}_{adv}$ is the adversarial loss function which we will define below. Intuitively, the first term performs parameter-efficient fine-tuning by leveraging the information present in the frozen backbone, while the second term promotes robustness.

\hfill \break
\noindent In this paper, we propose different choices for $\mathcal{L}_{adv}$:
\hfill \break

\paragraph{Cross Entropy (CE).} We can choose the cross-entropy loss for the prediction of the adaptive adversarial example:
\begin{align}
    \mathcal{L}_{adv} = \mathcal{L}_{CE} (f(\theta_p, x_p'), y)
\end{align}
which results in a loss function that promotes correct classification of samples perturbed with the adaptive attack.
\hfill \break

\paragraph{KL Divergence.} We can choose $\mathcal{L}_{adv}$ to be the KL-divergence between the prediction distributions of the unperturbed and perturbed samples: 
\begin{align}
    \mathcal{L}_{adv} = \mathcal{L}_{KL} (f(\theta_p, x_p'), f(\theta_p, x))
    \label{loss:KL}
\end{align}

With this formulation, the first term in \cref{loss:ADAPT} promotes the correct prediction of unperturbed samples, and the adversarial term \cref{loss:KL} matches the prediction distributions of unperturbed and perturbed samples.

\subsection{Additional Design Choices}
\label{sec:design}
\paragraph{Prompting the feature extractor.}
Prompt tuning can be performed in a multitude of ways by changing the definition of $f(\theta_p,x)$. In our experiments, we observe that for adversarial robustness, using PT2 (otherwise known as Prefix Tuning) significantly outperforms using only PT. This behavior is consistently observed when using the same training scheme and total number of tuned parameters across the two methods.
We hypothesize that this performance gap exists since the feature extractor $\Phi^*$ of the frozen backbone is not adversarially robust. Since PT simply prepends additional tokens only to the input of the frozen feature extractor, this results in the same function $\Phi^*$ on a different set of inputs. 
However, PT2 prepends prompt tokens directly in each layer of the feature extractor, resulting in a modified feature extractor $\hat{\Phi}$.

\paragraph{Tuning the embedding layer.}
Traditionally, the prompt tokens are introduced \textit{after} the embedding function $T$ and do not influence the mapping of the image patches to the embedding space.
That is, the patch embedding remains unchanged by prompt tuning, which can cause vulnerability. Since the patch embedding does not have a significantly large number of parameters compared to the frozen backbone, we opt to tune the patch embedding parameters $T$ along with the prompt tokens $\theta_p$ and the linear classifier head $\Psi$.

\Cref{alg:adapt} shows a pseudocode of our training algorithm. We provide our code as supplementary material, and we will release an open-source version of our code upon acceptance.

\begin{algorithm}
\caption{ADAPT training step}\label{alg:adapt}
\begin{algorithmic}
\Require data pair $x,y$, frozen classifier $f$, tunable prompts $\theta_p$, attack step size $\alpha$, maximum perturbation magnitude $\epsilon$, number of attack steps $s$, learning rate $\eta$
\State $x'_p \gets x + U(-\epsilon, \epsilon)$ \Comment{$U$ is the uniform noise distribution}
\For{$i = 1 \rightarrow s$}
\State $x'_p \gets x'_p + \alpha\nabla_{x'_p}\mathcal{L}_{CE} (f(\theta_p, x'_p), y)$ \Comment{\cref{eq:adpatk}}
\State $x'_p \gets$  Project($x'_p, x-\epsilon, x+\epsilon$) \Comment{Project $x'_p$ onto the $\epsilon$ ball}
\EndFor
\State $\mathcal{L} \gets \mathcal{L}_{ADAPT}$\Comment{\cref{loss:ADAPT}}
\State $\theta_p \gets$ $\theta_p + \eta\nabla_{\theta_p} \mathcal{L}$ \Comment{Optimizer gradient step}
\end{algorithmic}
\end{algorithm}

\section{Experiments}
\label{sec:experiments}

We evaluate our methods on benchmark datasets with a focus on increasing adversarial robustness while keeping the number of tuned parameters relatively low. When using prompt tuning, we demonstrate the gradient obfuscation occurring in previous methods, and show that our proposed method overcomes this issue. Finally, we perform ablation studies on the different components of our method. To the best of our knowledge, we are the first to explore the robustness of the prompt tuning paradigm for ViTs.

It is imperative to note that the advantage of the prompt tuning paradigm is the adaptation of a single frozen backbone to different tasks by using a small number of additional parameters per task. For example, if a model was required for a hundred different downstream tasks, traditionally we would have to save and load a hundred fully fine-tuned copies of the large backbone with hundreds of millions of parameters for each task. Instead, prompt tuning requires training only a fraction of the original parameters per task to adapt a single frozen pre-trained model for each downstream task. With that said, prompt tuning does not benefit the time required for training or inference for the downstream tasks as the prompts are used in conjunction with the entire frozen backbone. This is true for any method that adopts the prompt tuning paradigm and not just our scenario. In any case, we report training time details in the appendix.

\subsection{Experimental Setting}
\paragraph{Datasets.} We perform experiments on CIFAR-10\citep{cifar10}, CIFAR100 \citep{cifar100}, and Imagenette \citep{imagenette}. Note that these datasets are widely employed benchmarks for evaluating adversarial defenses since training and evaluation on ImageNet using multi-step gradient-based attacks requires significant computational time and resources.

\paragraph{Adversarial Robustness Methods.} We compare ADAPT to the following state-of-the-art methods established in the literature:

\begin{itemize}
    \item \textit{Adversarial Training }\citep{madry2018towards}: As described in \cref{eq:at}, Adversarial Training solves a minimax optimization objective by augmenting the inputs during training using traditional adversarial attacks. 
    \item \textit{TRADES} \citep{zhang2019theoretically}: TRADES identifies a trade-off between robustness and accuracy. It designs a defense method accordingly, which performs training with a modified loss function, performing natural training and at the same time reducing the prediction distance of unperturbed samples with that of samples perturbed to maximize the same distance.
    \item \textit{MART} \citep{wang2019improving}: MART builds on top of TRADES and emphasizes on misclassified examples during training to improve robustness.
    \item \textit{NFGSM} \citep{de2022make} NFGSM proposes to use a high-magnitude noise perturbation as an initialization point for generating adversarial attacks during single-step adversarial training to avoid overfitting
\end{itemize}


\paragraph{Architectures and Tuned Parameters.}
We perform most of our experiments on the ViT-B model. For prompt tuning (PT), we keep the number of tuned parameters within $\sim 1\%$ of the total model parameters. We provide additional results on the ViT-S and ViT-L model configurations.

As discussed in design choices, due to superior performance, we perform most experiments with PT2 and tune the patch embedding as well as the linear classifier (all of which we take into account into the number of tuned parameters). We also perform ablation studies on PT while freezing the patch embedding layer.

\paragraph{Evaluation Metrics.} We evaluate each method with regards to accuracy on the test set. We evaluate on unperturbed samples, also known as clean accuracy, as well as perturbed samples. In one experiment, as mentioned, we use traditional adversarial attacks as well as adaptive ones to validate our claims of gradient obfuscation. In all other sections, excluding the black-box attacks section, our attacks are white-box and adaptive to ensure a fair worst-case evaluation.  We use adaptive PGD10 \cref{eq:adpatk}, adaptive CW \citep{carlini2017towards}, which is obtained by replacing the loss in \cref{eq:adpatk} with the CW loss, as well as AutoAttack (AA) \citep{croce2020reliable}, an ensemble of diverse and parameter free white-box and black-box attacks. Finally, we evaluate against two state of the art blackbox attacks: RayS \citep{chen2020rays} and AdaEA \citep{chen2023adaptive}.

\subsection{False Sense of Security} \label{exp:false}
In this section, we evaluate the robustness of existing methods, combined with prompt tuning, against traditional and adaptive attacks. The results are presented in \cref{tab:napt} and are consistent with what we discussed in adaptive attacks section. The results show that using traditional adversarial examples and adversarial training can lead to {gradient obfuscation} \citep{athalye2018obfuscated}, and a false sense of security. Indeed, existing adversarial defense methods initially exhibit high performance when used with prompt tuning. However, as per \citep{athalye2018obfuscated}, a sign of {gradient obfuscation} is that single-step gradient-based attacks are more effective than their multi-step counterparts. As expected, previous methods are more susceptible to single-step FGSM attacks than multi-step PGD attacks and show significant vulnerability to our adaptive attack. 

\begin{table}
\small
  \centering
  \footnotesize
  \caption{Test set accuracy of previous adversarial robustness methods combined with prompt tuning against unperturbed inputs, traditional adversarial attacks, PGD10 (multi-step) and FGSM (single step) as well as our adaptive attack. All methods perform worse against FGSM than PGD10, a clear sign of gradient obfuscation. Additionally, all methods show very little robustness to the adaptive attack (Adaptive PGD10). $^\dagger$ indicates that the attack in question is non-adaptive while $^*$ indicates an adaptive attack \cref{eq:adpatk}. }
  \begin{tabular}{c |c| c| c| c| c}
    \toprule
    Params & Method & Clean & PGD$^\dagger$ & FGSM$^\dagger$ & PGD$^*$\\
    \midrule
    PT & AT & 93.21 & 87.62 &  78.92 & 1.15\\
    PT2 + Emb& AT & 94.84 & 90.3 & 82.49 & 3.04 \\
    PT & TRADES & 89.79 & 84.87	& 79.65 & 1.01 \\
    PT2 + Emb &TRADES &92.64& 89.65  & 85.93 & 3.17 \\
    \bottomrule
  \end{tabular}
  
  \label{tab:napt}
\end{table}

\subsection{Adaptive Adversarial Training}
\label{exp:adapt}
We experiment with prompt tuning using previous SOTA adversarial defense methods and compare them to ADAPT. We present the results in \cref{tab:apt}. The results show that while previous methods show little to no robustness against adaptive adversarial attacks, ADAPT achieves competitive accuracy on unperturbed samples while showing considerable robustness to adaptive adversarial attacks. We provide additional results for different model sizes, ViT-L and ViT-S, in the appendix, the results are consistent with that for ViT-B.

We also evaluate our methods on Imagenette \citep{imagenette}, which is a subset of ImageNet with 10 classes, and CIFAR100 \citep{cifar100}, which consists of 100 classes of 32 $\times$ 32 images. Results are presented in \cref{tab:imagenette} and \cref{tab:cifar100} respectively, and show consistency with the CIFAR10 experiments. 

Note that for a fair comparison, we perform PT2 with the same number of tokens and tune the patch embedding and the linear classifier head for all methods, resulting in exactly the same number of tuned parameters. We include the results of full model fine-tuning (FT) as an upper bound on performance, which requires tuning all the ViT parameters. We conduct experimentation with PT and the effects of training the patch embedding in the ablation study (\cref{tab:ablation}). The results confirm the effectiveness of our design choices. Furthermore, we investigate the impact of the number of tuned parameters in the following sections.

\begin{table}[h!]
\small
  \centering
  \setlength{\tabcolsep}{1mm}
  \caption{Test set accuracy on the CIFAR10 dataset for different methods. The upper section corresponds to methods using prompt tuning, and the lower section corresponds to FT as an upper bound on performance. Existing methods show little to no robustness to adaptive attacks. In the prompt tuning scenario, our method shows substantial improvement in robustness compared to previous methods. $^*$ indicates that the attack in question is adaptive (\cref{eq:adpatk}).}
  \begin{tabular}{c| c| c| c| c| c| c}
    \toprule
     Method &Params.  &  Clean & PGD10$^*$ & CW$^*$ & AA & \# params\\
    \midrule

     AT &PT2+Emb & \textbf{94.84} & 3.04 &  0.67&  1.7 &  820K  \\
    TRADES &PT2+Emb &   92.64 & 3.17 & 1.85 & 4.2 & 820K \\
    MART & PT2+Emb& 52.28 & 12.1 & 9.62 & 15.25 &820K \\
    NFGSM & PT2+Emb & 77.95 & 0.1 & 0.6 & 2.03& 820K \\
    \rowcolor{lightgray}$\text{ADAPT}_{CE}$ &PT2+Emb & {79.05}&	38.27 & \textbf{35.94}	 & \textbf{19.90} & 820K \\
    \rowcolor{lightgray}$\text{ADAPT}_{KL}$  &PT2+Emb &  68.43 & \textbf{40.21} & 35.9 & 18.05 &  820K \\
    \midrule
    AT &FT &   {82.42} & {53.41} & {46.67}  & 34.77 &  86M \\ 
    TRADES &FT  & \textbf{84.02} & \textbf{54.19} & \textbf{47.45}  & \textbf{36.48} &  86M \\
    \bottomrule
  \end{tabular}
  
  \label{tab:apt}
\end{table}

\begin{table}[h!]
\footnotesize
    \centering
    \caption{Test set accuracy on Imagenette}
    \begin{tabular}{c| c| c| c| c |c|c}
    \toprule
         Method & Params &Clean & PGD10 & CW & AA& \# params \\
         \midrule
         TRADES & PT2+Emb & \textbf{89.32} & 9.04 & 6.82& 7.08&820K\\
         MART & PT2+Emb&44.2 & 17.35 & 14.03 & \textbf{19.36}&820K\\
         NFGSM & PT2+Emb & 71.82 & 2.08 & 2.08 & 5.68 & 820K \\
         \rowcolor{lightgray}$\text{ADAPT}_{CE}$ & PT2+Emb&59.95 & 31.34	& 28.25 & {13.81} &820K \\
         \rowcolor{lightgray}$\text{ADAPT}_{KL}$ & PT2+Emb&63.16& \textbf{37.61} & \textbf{34.01} & 10.96& 820K \\
         \midrule
         {TRADES}& FT &\textbf{82.4} & \textbf{53.4} & \textbf{43.6} & \textbf{16.33}&86M\\
         \bottomrule
    \end{tabular}
    
    \label{tab:imagenette}
\end{table}

\begin{table}[h!]
\footnotesize
    \centering
    \caption{Test set accuracy on CIFAR100}
    \begin{tabular}{c |c |c |c |c|c|c}
    \toprule
         Method & Params&Clean  & PGD10 & CW& AA&\# params \\
         \midrule
         TRADES & PT2+Emb & \textbf{78.84} & 5.54 & 3.52 & 3.4&820K\\
         MART & PT2+Emb &29.29 & 6.59 & 4.99 & \textbf{8.5} & 820K\\
         NFGSM & PT2+Emb & 57.91 & 0.6& 0.9 & 1.67& 820K  \\
         \rowcolor{lightgray}$\text{ADAPT}_{CE}$ & PT2+Emb &60.2 & \textbf{22.15}	&\textbf{19.59} & {7.92}& 820K\\
         \rowcolor{lightgray}$\text{ADAPT}_{KL}$ & PT2+Emb &50.63 & 21.77 & 17.56& 5.94& 820K\\
         \midrule
         {TRADES} & FT& \textbf{65.5}&	\textbf{32.89}  & \textbf{28.93} & \textbf{9.99} & 86M\\
         \bottomrule
    \end{tabular}
    
    \label{tab:cifar100}
\end{table}

\subsection{Evaluation against Blackbox Attacks}
In this section, we evaluate the robustness of ADAPT and existing methods against black-box attacks in the prompt-tuning paradigm. We evaluate against two SOTA black-box attacks: one query-based attack, RayS \citep{chen2020rays} and one transfer-based ensemble attack, AdaEA \citep{chen2023adaptive}.
\Cref{tab:bb} shows that ADAPT is robust to \textbf{both} the query-based and transfer-based attacks, while AT and TRADES (with prompt tuning) only show robustness to the transfer-based attack and are \textbf{not robust} to the query-based attack. We hypothesize that the reason is AT and TRADES (with prompt tuning) are trained on attacks generated using the gradient of the naturally trained frozen ViT. The transfer-based attack also relies on the gradients of naturally pre-trained models to construct its attack, while the query-based attack eliminates the need to use gradients. This aligns with our discussion of gradient obfuscation. Furthermore, against existing defenses, both black-box attacks are {significantly} more successful than white-box non-adaptive {attacks} {(see \cref{tab:apt})}, which is another sign of gradient obfuscation {in existing defenses when using prompt tuning}.
 
\begin{table}[h!]
    \footnotesize
    \centering
    \caption{Test-set accuracy against black-box attacks, same training settings as Table 2.}
    \begin{tabular}{c|c|c|c}
    \toprule
    Method &Params& RayS  & AdaEA  \\
    \midrule
    AT     &P2T+Emb&  5.06 &66.55\\
    TRADES     &P2T+Emb& 11.71 &66.27\\
    \rowcolor{lightgray}$\text{ADAPT}_{CE}$ &P2T+Emb& 43.05 & 58.21\\
    \rowcolor{lightgray}$\text{ADAPT}_{KL}$ &P2T+Emb& 40.09& 52.06\\
    \midrule
    TRADES & FT & 56.78 & 61.73 \\
    \bottomrule
    \end{tabular}
    
    \label{tab:bb}
\end{table}

\subsection{Beyond the Number of Parameters}

We notice in our experiments that for adversarial training under the same conditions, PT2 significantly outperforms PT. We hypothesized that this is due to the fact that PT is simply augmenting the input to the frozen feature extractor $\Phi*$ of the transformer which is vulnerable to adversarial examples, while PT2 inserts tokens in the intermediate layers of the feature extractor, modifying it directly. We perform an experiment to validate our intuition and confirm that the superior performance of PT2 over PT, is not simply due to an increased number of parameters (as with a fixed number of tokens per layer, PT2 requires more tuned parameters).


We consider a set of experiments with a fixed number of tuned tokens, and we gradually change the scenario from PT to PT2 in discrete steps as described below. We tune a fixed number of $M$ tokens, prompting the first $k$ layers of the transformer with $m$ tokens each, with varying numbers of $k$ and $m$ while keeping $m\times k = M$. 

For example, we consider a pool of $96$ tokens. In the first experiment, we perform training and testing while only prompting the first layer with all $96$ tokens (which is equivalent to PT). In the following experiments, we perform training and testing while prepending to more and more layers, e.g. the first two layers with $48$ tokens, and so on, until prepending to all $12$ layers each with $8$ tokens (which is equivalent to PT2). In all of the above experiments, the same number of tuned parameters is used ($96$ tokens, plus the classifier head). 

The results are presented in \cref{tab:lendep}, and show a substantial increase in performance as we prompt more and more layers despite using the same number of tokens. This further validates our insight about needing to modify the feature extractor and shows that the performance improvement is not simply due to an increase in the number of tuned parameters.
\begin{table}[h!]
\small
    \centering
    \caption{Test set accuracy and the number of tuned parameters of different combinations. $m\times k$ means we prompt the first $k$ layers directly with $m$ tokens each. Each setting is adopted during both training and testing. We measure against unperturbed samples and the adaptive PGD10 attack.}
    \begin{tabular}{c|c|c|c}
    \toprule
    Combination & Clean  & PGD10 & \# params\\
    \midrule
        96$\times$1 & 27.33 &18.34 & 81K \\
        48$\times$2 & 32.74 &21.07 & 81K\\
        24$\times$4 & 40.03 & 24.83 & 81K\\
        16$\times$6 & 40.78 & 25.49 & 81K\\
        12$\times$8 & \textbf{41.14} & \textbf{25.68} & 81K\\
        8$\times$12 & 40.24 & 25.17 & 81K\\
    \bottomrule
    \end{tabular}
    
    \label{tab:lendep}
\end{table}

\begin{table}[h!]
\footnotesize
    \centering
    \caption{Ablation study on different prompt tuning scenarios on the CIFAR10 dataset. Performance is measured by test set accuracy againt unperturbed samples and samples perturbed by adaptive PGD10.}
    \begin{tabular}{c| c| c| c| c}
    \toprule
         Method & Params & Clean & PGD10 & \# params \\
         \midrule
         $\text{ADAPT}_{CE}$ & PT  & 44.72 & 15.24 & 19K\\
         $\text{ADAPT}_{KL}$ & PT  & 39.77& 9.54& 19K\\
         \midrule
         $\text{ADAPT}_{CE}$ & PT2  & 67.45& 26.75& 230K\\
         $\text{ADAPT}_{KL}$ & PT2  & 58.4 & 29.25& 230K\\
         \midrule
         $\text{ADAPT}_{CE}$ & PT+Emb  & {67.8} & 29.44 & 600K\\
         $\text{ADAPT}_{KL}$ & PT+Emb & 56.48 & {30.31} & 600K\\
        \midrule
        $\text{ADAPT}_{CE}$ &PT2+Emb & \textbf{79.05}& 38.27 &  820K \\
        $\text{ADAPT}_{KL}$ &PT2+Emb &  67.24 & \textbf{39.23} & 820K \\
         \bottomrule
    \end{tabular}
    
    \label{tab:ablation}
\end{table}

\section{Conclusion}
We formulated and extensively analyzed the adversarial robustness of ViTs while performing parameter-efficient prompt tuning. We found that the application of existing adversarial defense methods to the prompt tuning paradigm suffers from gradient obfuscation. This leads to a \textit{false sense of security} and a vulnerability to adaptive attacks, which we demonstrated through experiments with single-step and adaptive attacks.
To address these problems, we proposed a novel loss function for adaptive adversarial training of prompt tuning methods. 
We evaluated our method on various datasets and model sizes, and performed an ablation study on the components of our framework.
We empirically demonstrated the superior adversarial robustness of our method while performing parameter-efficient prompt tuning.

\section{Broader Impact Statement}
Adversarial perturbations can be used to fool machine learning models to make decisions based on the adversary's intent which can be malicious. For example, in facial recognition for security systems, an adversary may try to pose as someone else to breach the system. In self-driving cars, an adversary can change the classification of traffic signs or objects to alter the course of the car and cause deadly crashes.
As such, it is imperative to explore and improve the robustness of machine learning models. 
In this work, we showcase the significant vulnerability of previous adversarial defense methods to adaptive adversarial attacks under the prompt tuning paradigm. To this end, we provide a framework for adaptive adversarial training of prompts to combat vulnerability to adaptive attacks. This makes models using the prompt tuning paradigm more robust to adversarial manipulation. 

\section{Acknowledgments}

This project was supported by NIH grants R01CA240771 and U24CA264369 from NCI. 
We would like to thank the TMLR editors and reviewers as well as our colleagues and friends for helping improve this work.

\bibliography{main}
\bibliographystyle{tmlr}

\appendix
\section{Different Model Sizes}
We provide results for the experiment setting of table 1, for different ViT configurations, namely ViT-S in \cref{tab:vits} and ViT-L in \cref{tab:vitl}. The results are consistent with that of table 1.

\begin{table}[h!]
    \footnotesize
    \centering
    \caption{CIFAR10 test set accuracy for the ViT/Small model}
    \begin{tabular}{c| c| c| c| c| c}
    \toprule
         Method & Params. &Clean & PGD & CW & \#  params \\
         \midrule
         TRADES & PT2+Emb& \textbf{86.06}& 3.98& 3.81& 740K\\
         ADAPT+CE &PT2 + Emb & 73.7& 34.33& {31.52} & 740K\\
         ADAPT+KL & PT2 + Emb & 63.37 & \textbf{36.1}& \textbf{31.74}& 740K\\
         \midrule
         {TRADES} & FT&\textbf{81.1}&	 \textbf{51.71} & \textbf{45.16}& 57M \\
         \bottomrule
    \end{tabular}
    \label{tab:vits}
\end{table}

\begin{table}[h!]
    \footnotesize
    \centering
    \caption{CIFAR10 test set accuracy for the ViT/Large model. }
    \begin{tabular}{c| c| c| c| c|c}
    \toprule
         Method & Params. &Clean  & PGD & CW & \#  params\\
         \midrule
         TRADES & PT2+Emb& \textbf{94.12}& 3.68	& 2.36&1.4M\\
         ADAPT+CE & PT2+Emb&80.79&	39.49& \textbf{36.82} &1.4M	\\
         ADAPT+KL & PT2+Emb&69.98&  \textbf{41.16} & \textbf{36.82}&1.4M \\
         \midrule
         {TRADES} & FT& \textbf{83.88}&	 \textbf{54.67} & \textbf{48.12} & 307M\\
         \bottomrule
    \end{tabular}
    \label{tab:vitl}
\end{table}

\section{Training Time Analysis}
\label{app:time}
Prompt tuning seeks to adapt large models to downstream tasks with minimal parameters. This allows us to load and store only one set of weights for a large backbone, and that set can be adapted by storing and loading a low number of parameters for each downstream task.

Subsequently, it is important to note that the advantage of prompt tuning is a reduction in the number of tuned parameters and not a reduction in training time, as the forward and backward passes still flow through the entire model. This results in a slight increase in computation cost, as adding prompt tokens increases the number of tokens that go through the transformer. \textit{This is the case for any application of PT and PT2 and is \textbf{not} specific to our scenario}. However, in our experiments, we empirically observe that while tuning the prompts, we can train for fewer epochs with a cyclic learning rate to achieve our best-case performance. The same was not observed in the fine-tuning scenario and more training epochs were required to achieve the best performance in terms of validation accuracy.

With all that said, we provide training time details for the scenarios in \cref{tab:traintime} for those interested. We report the training time for each method in minutes. Each method was trained using an NVIDIA Tesla V100 SXM2.

\begin{table}[h]
\small
  \centering
  \caption{Total training time as measured in hours (rounded to the closest integer) for the methods presented in Tab. 2 of the main paper. Each method in the prompt tuning scenario was trained for 20 epochs with a cyclic learning rate while each method in the fine tuning scenario was trained for 40 epochs with an annealing learning rate.}
  \begin{tabular}{c| c| c}
    \toprule
     Method &Params.  &  Total Training Time (hours)\\
    \midrule

     AT &PT2+Emb & 15  \\
    TRADES &PT2+Emb &   16  \\
    ADAPT-CE (Ours) &PT2+Emb & 18 \\
    ADAPT-KL (Ours) &PT2+Emb &  18\\
    \midrule
    AT &FT &   11 \\ 
    TRADES &FT  & 13 \\
    \bottomrule
  \end{tabular}
  
  \label{tab:traintime}
\end{table}
\end{document}

%% file: math_commands.tex
\usepackage{amsmath,amsfonts,bm}

\def\eqref#1{equation~\ref{#1}}

\def\1{\bm{1}}

\DeclareMathAlphabet{\mathsfit}{\encodingdefault}{\sfdefault}{m}{sl}
\SetMathAlphabet{\mathsfit}{bold}{\encodingdefault}{\sfdefault}{bx}{n}

